\documentclass[letterpaper, 10 pt, conference]{ieeeconf}

\usepackage{times}

\usepackage{graphics} 
\usepackage{epsfig} 
\usepackage{mathptmx} 
\usepackage{times} 

\usepackage{amsmath} 
\usepackage{amssymb}  
\usepackage{float}
\usepackage[labelformat=simple]{subcaption}

\usepackage{bm}

\usepackage{comment}

\usepackage[utf8]{inputenc} 
\usepackage[T1]{fontenc}    
\usepackage{hyperref}       
\usepackage{url}            

\usepackage{booktabs}       
\usepackage{amsfonts}       
\usepackage{nicefrac}       
\usepackage{microtype}      
\usepackage{mathtools}

\usepackage{amsfonts}
\usepackage{amsthm}
\usepackage{amsmath}
\usepackage{amssymb}

\usepackage[noend]{algorithm,algpseudocode}
\usepackage{wrapfig}

\newtheorem{example}{Example}

\theoremstyle{definition}

\newtheorem{definition}{Definition}

\newcommand{\lie}{\mathsf{L}_{f}}
\newcommand{\dlie}{\mathsf{L}_{f,\Delta t}}

\DeclareMathOperator*{\argmin}{arg\,min}

\algnewcommand{\IIf}[1]{\State\algorithmicif\ #1\ \algorithmicthen}
\algnewcommand{\EndIIf}{\unskip\ \algorithmicend\ \algorithmicif}

\DeclareMathAlphabet{\mathcal}{OMS}{cmsy}{m}{n}

\newcommand{\fixed}[2]{#1}

\title{Quantifying Safety of Learning-based Self-Driving Control Using Almost-Barrier Functions}
\author{\textbf{Zhizhen Qin, Tsui-Wei Weng, Sicun Gao}\\\textnormal{University of California San Diego}\\\texttt{\{zhizhenqin, lweng, sicung\}@ucsd.edu}}

\begin{document}
\bstctlcite{IEEEexample:BSTcontrol}
\maketitle

\begin{abstract}
Path-tracking control of self-driving vehicles can benefit from deep learning for tackling longstanding challenges such as nonlinearity and uncertainty. However, deep neural controllers lack safety guarantees, restricting their practical use. We propose a new approach of learning almost-barrier functions, which approximately characterizes the forward invariant set for the system under neural controllers, to quantitatively analyze the safety of deep neural controllers for path-tracking. We design sampling-based learning procedures for constructing candidate neural barrier functions, and certification procedures that utilize robustness analysis for neural networks to identify regions where the barrier conditions are fully satisfied. We use an adversarial training loop between learning and certification to optimize the almost-barrier functions. The learned barrier can also be used to construct online safety monitors through reachability analysis. We demonstrate effectiveness of our methods in quantifying safety of neural controllers in various simulation environments, ranging from simple kinematic models to the TORCS simulator with high-fidelity vehicle dynamics simulation.
\end{abstract}
\section{Introduction}

Safe path-tracking control is crucial for reliable autonomous driving. Widely-adopted control methods~\cite{Snider-2009-10165,drivingsurvey} have inherent difficulty with nonlinearity and uncertainty that cannot be ignored when the vehicles are operating at relatively high speed or under adverse road conditions. Controllers obtained from deep learning methods have shown great promise in a variety of application \cite{DBLP:conf/rss/KaufmannLR0K020,DBLP:conf/rss/PengCZLTL20}. However, 
neural networks are known to be highly nonlinear and complex, preventing them from being easily analyzed as classical controllers such as Stanley~\cite{DBLP:journals/jfr/ThrunMDSADFGHHLOPPSSDJKMRNJABDEKNM06} or Model Predictive Control (MPC)~\cite{drivingsurvey,DBLP:conf/ivs/KongPSB15}.
In this paper, we propose methods for the quantitative safety analysis of learning-based neural controllers by synthesizing and certifying {\em neural almost-barrier functions}, for path-tracking with only black-box access to high-fidelity simulations of nonlinear vehicle dynamics.

Barrier functions~\cite{prajna2007framework,ames2014control,ames2017control} of dynamical systems can identify forward invariant sets of the system, which ensures that any trajectory of the system that starts within the invariant sets can never reach states that are outside the invariant sets. Such properties can be used to certify safety properties, by showing that the unsafe regions of the system are disjoint from the forward invariants. For instance, in the context of path-tracking control, an ideal forward invariant set forms a tube around the path that the vehicle is supposed to follow. If such invariant sets can be established, we know that the vehicle will stay within the tube around the path and avoid unsafe behaviors such as drifting off the path. 

An almost-barrier function is a relaxation of the standard notion of barrier certificates by allowing small regions on the safety barrier to be uncertified. This relaxation is critical for using invariant-based analysis for complex systems under neural network controllers: it allows us to certify safety for the majority of the state space, while clearly identifying small regions that are potentially unsafe.  
Such analysis is more informative than the standard barrier function that provides all-or-nothing results and is most suitable for systems with high nonlinearity that are typically beyond the scope of standard barrier function methods. The almost-barrier can also be used to construct online safety monitors that can alert human operators to take over ahead-of-time, only when the ``leaking" regions of the safety barrier may be reached. We believe the study of this relaxed notion of safety analysis is a suitable step for developing data-driven and model-free methods for highly complex control systems in practice. 

We will describe the sampling-based learning procedures for constructing candidate neural barrier functions, and certification procedures that utilize robustness analysis for neural networks to certify regions where the barrier conditions are fully satisfied. Our approach is built on recent advances in robustness certification of neural networks~\cite{weng2018towards,weng2019proven}, which allows us to rigorously bound the Lie derivative values of the learned neural barrier function. With these methods, we are able to train and certify barrier functions with small region of boundary violations: $99\%$ of the barrier region is fully certified for the kinematic vehicle model, 86\% for the highly nonlinear dynamic model with inertial effects and lateral slip, and 91\% in the TORCS environment with high-fidelity vehicle dynamics simulation (Fig.1). We visualize the certified regions (in blue contour) and the sparse uncertified regions (in red contour) of the learned safety barriers, and demonstrate how to construct online safety monitor through reachability analysis. 

Overall, we make the following contributions:

\begin{figure}[t]
\centering
\begin{subfigure}{.3\textwidth}
  \centering
  \includegraphics[width=\linewidth]{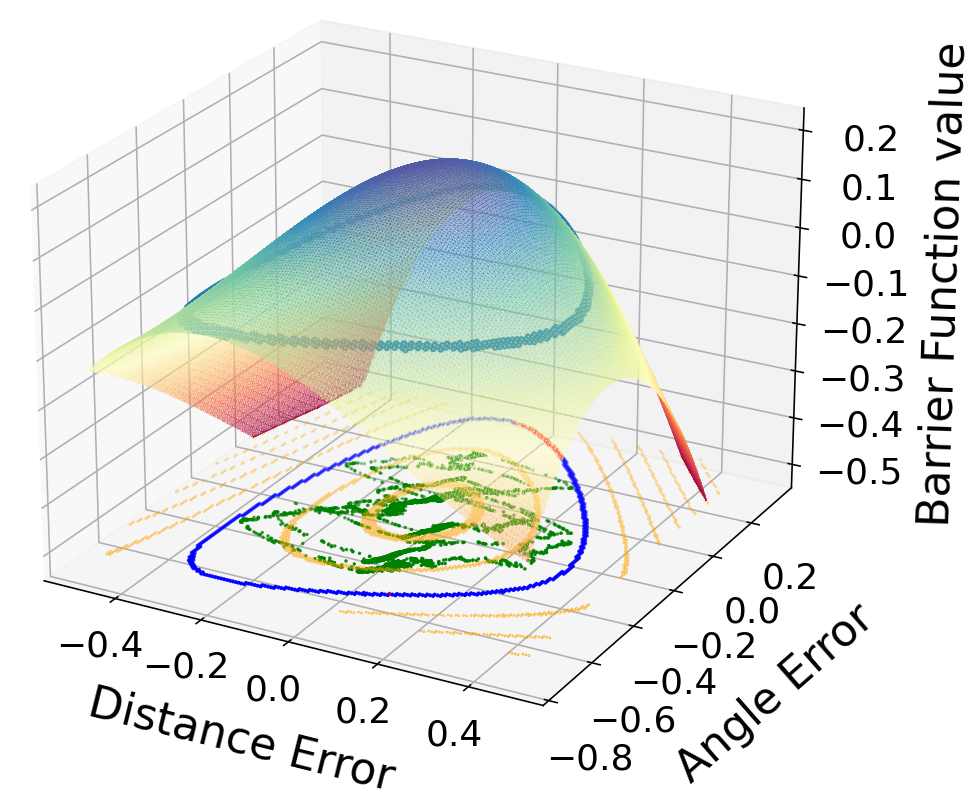}  
  \caption{}
  \label{fig:cbf_2d_torcs}
\end{subfigure}
\begin{subfigure}{.15\textwidth}
  \centering
  \includegraphics[width=\linewidth]{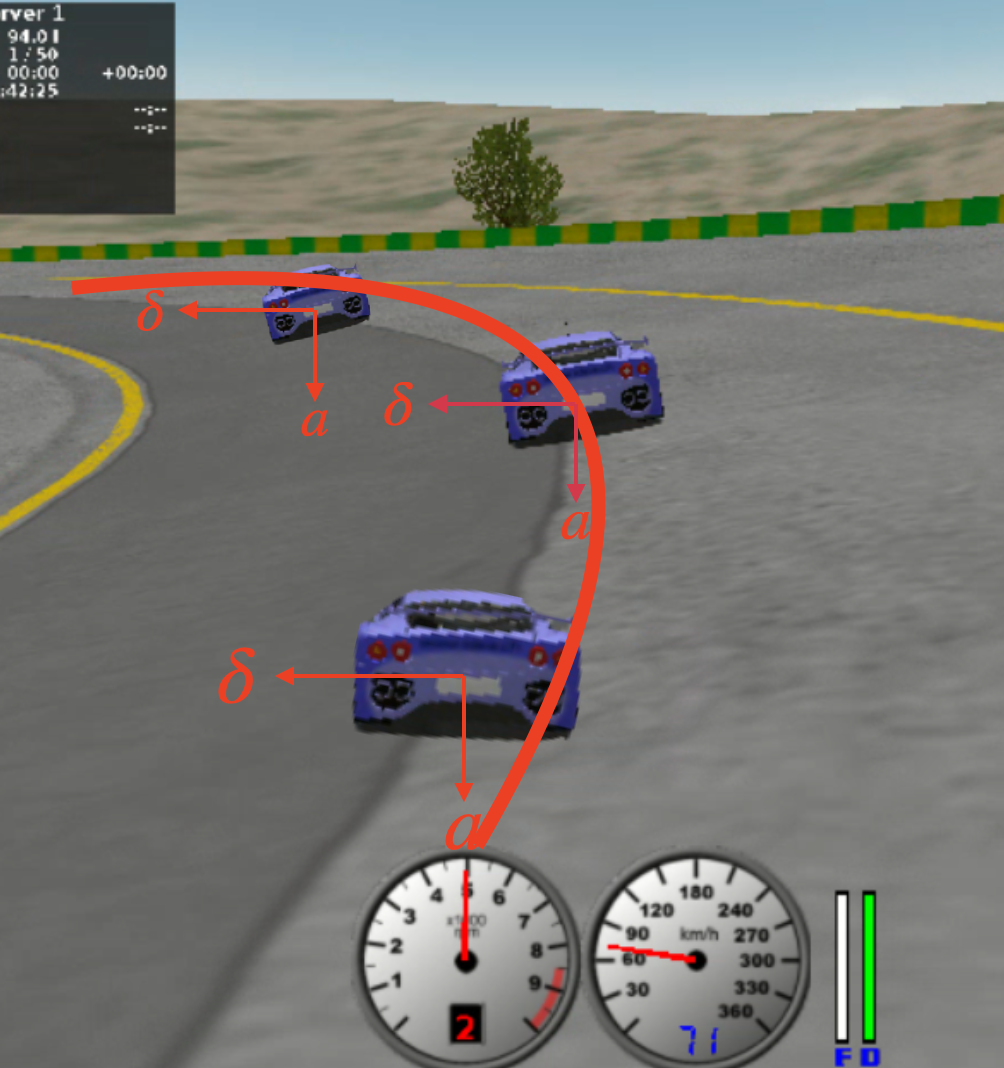}  
  \caption{}
  \label{fig:torcs_behavior}
\end{subfigure}
\caption{Learned neural almost-barrier function for deep neural path-tracking control in TORCS simulation environment. (a) The surface plot shows the learned almost-barrier function. The blue curve in x-y plane is the zero-level set, i.e., the safety boundary that separates the safe trajectories from unsafe states, with small portion of red curve identifying uncertified regions. (b) shows the TORCS simulation of the vehicle behavior near the safety boundary, where the neural control applies brake and steers the vehicle back to the safety.}\vspace{-0.5cm}
\label{fig:cbf_2d}
\end{figure}

\noindent \textbf{\#1}. We propose a systematic methodology to learn neural almost-barrier functions for learning-based neural controllers for path-tracking with either full states or partial observability (Section \ref{sec:learn_barrier_function}). 

\noindent \textbf{\#2}. We propose certification procedures that rigorously bound the Lie derivatives of the candidate neural barrier functions through sampling-based analysis (Section \ref{sec:certify_barrier_function}) . 

\noindent \textbf{\#3}. We evaluate the proposed methods using three types of simulation environments with increasing difficulty. We show quantitative certification results for the neural controllers in all environments (Section \ref{sec:exp}).

Te proposed method could be applied to arbitrary controllers, and we demonstrate it on neural controllers because they are intrinsically complex and hard to be analyzed. To the best of our knowledge, the proposed approach is the first to attempt quantitative safety analysis of deep neural controllers for path-tracking under nonlinear high-fidelity vehicle dynamics.

\noindent{\bf Assumptions.} We make the following assumptions for the proposed training and certifying procedures. 

\noindent{- Assumption 1:} The underlying dynamics of the vehicle is Lipschitz-continuous. To certify the barriers, we use numerical analysis procedures that assume the system dynamics is at least locally Lipschitz-continuous around each sample.
    
\noindent{- Assumption 2: } We assume capacity for drawing samples around the safety barrier (zero-level set of the learned barrier function). Successful certification requires accurate Lie derivative estimation around the boundary of the learned barrier. To ensure such requirement, we must be able to sample sufficient state transition data around the safety barrier to such that an estimate could be established. 
    
\noindent{- Assumption 3:} The trained control policy is close to being safe in the sense that there exists an almost-barrier function for the given controller. Otherwise the adversarial retraining loop may not converge.
    
\noindent{- Assumption 4:} The barrier function lies in the hypothesis class captured by neural networks. We use neural networks to represent the safety barrier and are thus assuming that the expressiveness of the neural network is enough for capturing the safety region. In practice, we found that if the safety region is of a highly non-convex shape, it becomes challenging to train a barrier function that describes the safe region.

\section{Related Work}
\paragraph{Safe Control for Autonomous Driving}  Certification of safety of path tracking control has been widely studied through reachability analysis~\cite{10.1007/11730637_22,epsilon-approximationof,10.1007/3-540-36580-X_5,1656431,4738704,DBLP:conf/hybrid/DangMT10,1166525,1215682}. Such methods guarantees safety by certifying reach-avoid sets. 
\cite{1166525,1215682} reformulates Hamilton-Jacobi equations to make the computational complexity only be exponential to state variables. \cite{8263867} uses simple dynamics models to bound tracking errors and then fine-tune the planning behavior close to obstacles to guarantee overall safety. 
The Stanley controller is a well-studied feedback control law for path-tracking in self-driving cars~\cite{DBLP:journals/jfr/ThrunMDSADFGHHLOPPSSDJKMRNJABDEKNM06,4282788,Snider-2009-10165,8998094,7795743}. It has exponential stability properties under low speed and tire no-slip assumption for the kinematic bicycle model, but the performance degrades in more challenging scenarios~\cite{DBLP:journals/jfr/ThrunMDSADFGHHLOPPSSDJKMRNJABDEKNM06}. 
Model-Predictive Control (MPC) methods~\cite{drivingsurvey,DBLP:conf/ivs/KongPSB15} are typically used for handling the highly nonlinear cases, but it requires high-frequency online optimization, precise dynamics modeling, and good initialization through human intervention~\cite{DBLP:conf/icra/SchwartingAPKR17}, while guaranteeing its safety is also challenging in nonlinear cases. 
\paragraph{Control Barrier Functions} Barrier certificates~\cite{prajna2007framework} and control barrier functions \cite{wieland2007constructive,ames2014control,ames2017control,parrilo2000structured} were proposed for certifying the safety of controlled dynamical systems by guaranteeing that all states of the system are contained within a forward invariant set. 
Synthesizing barrier functions for general nonlinear systems is a well-known challenge with various analytic and optimized-based approaches for systems with known dynamics~\cite{ames2019control,xu2017correctness,wang2018permissive}. The work in \cite{DBLP:conf/aaai/ChengOMB19} uses control barrier functions to ensure safety of learned control policies. The work in \cite{DBLP:conf/icml/ChengVOCYB19} proves stability properties throughout learning by taking advantage of the robustness of control-theoretic priors. Learning-based approaches for constructing barrier functions~\cite{DBLP:conf/cdc/RobeyHLZDTM20,DBLP:journals/fac/ZhaoZCLW21} have recently been proposed for systems with known dynamics. 
\paragraph{Robustness verification of neural networks} 
As deep neural networks (DNN) become prevalent in machine learning their vulnerability to adversarial examples motivated the need of quantifying robustness of DNN models~\cite{katz2017reluplex}. 
Recent verification methods based on local Lipschitz constant and convex relaxation have been proposed to compute the best lower bounds of the maximum robustness certificate~\cite{hein2017formal,weng2018evaluating,kolter2017provable,weng2018towards,zhang2018crown,singh2018fast,wang2018efficient,raghunathan2018semidefinite,Boopathy2019cnncert,ko2019popqorn,weng2019proven}. We will primarily adapt the efficient linear bounding framework in \cite{weng2018towards,zhang2018crown,Boopathy2019cnncert} to certify the neural almost-barrier functions.
\begin{figure*}[ht!]
    \centering
    \includegraphics[width=0.3\textwidth]{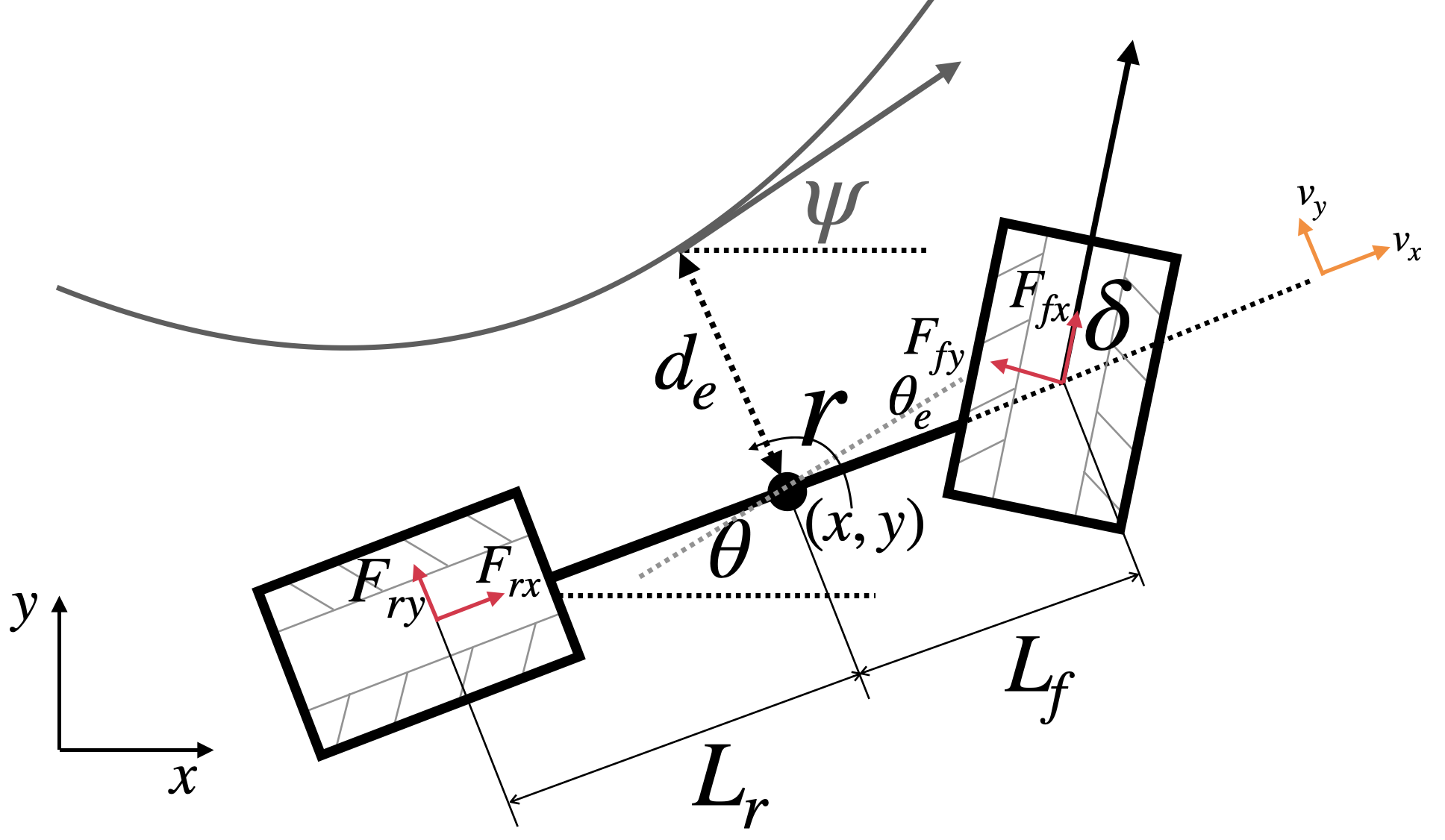}
    \includegraphics[width=0.6\textwidth]{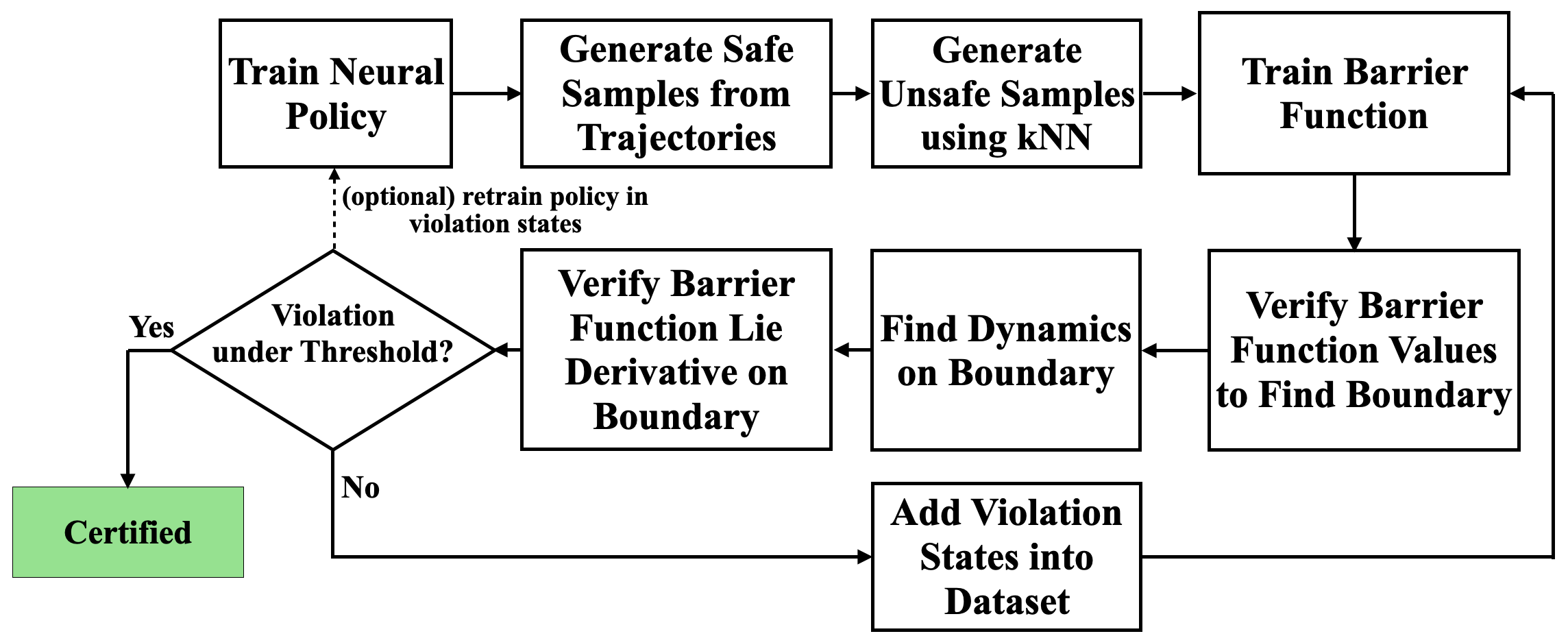}
    \caption{Left: Vehicle dynamics with lateral slip. Right: Overall flowchart for learning and certifying almost-barrier functions.}\vspace{-0.5cm}
    \label{fig:bicycle_models_and_flow}
\end{figure*}

\section{Preliminaries}
We consider controlled dynamical systems defined as
\begin{equation}
\label{eqn:dynamics}
    \dot{x}(t)=f(x(t), u(t)), \; u(t)=g(x(t)), \; x(0)=x_{0}
\end{equation}
where $x(t)$ is an $n$-dimensional state vector taking values in a state space $X\subseteq \mathbb R^n$, $u(t)$ is an $m$-dimensional control vector, and $f:X\times \mathbb{R}^{m} \rightarrow \mathbb{R}^{n}$ is a Lipschitz-continuous vector field. We focus on systems with an explicit control law $g:X\rightarrow \mathbb{R}^m$. 
Given such as system, we can define an unsafe region of the state space $X_u\subseteq X$ and say the system is safe if its trajectories never intersect with $X_u$. Formally, this means that there exists a {\em forward invariant set} for the system that is disjoint from the unsafe set. We say $I\subseteq X$ is a forward invariant set for the system in (\ref{eqn:dynamics}) if for any $x(0)\in I$ and any $t\geq 0$, we have $x(t)\in I$. Namely, any trajectory that starts in the invariant $I$ stays in $I$ forever. Consequently, a system is safe if we can find an invariant $I$ such that $I\cap X_u=\emptyset$. 

Barrier functions are important for establishing forward invariant sets. The idea is to use a scalar function to represent the boundary of a set, and show that the controlled dynamics of the system can never cross this boundary and escape the set. Formally we need the concept of the Lie derivative of scalar functions in vector fields. Without loss of generality we assume the system is time-invariant and omit $t$.
\begin{definition}[Lie Derivatives]\label{def:lie} 
Consider the system in (\ref{eqn:dynamics}) and let $B:X\rightarrow \mathbb{R}$ be a continuously differentiable function. The Lie derivative of $B$ over $f$ is defined as \begin{equation} \label{eqn:lie}
    \lie B(x)=\sum_{i=1}^{n} \frac{\partial B}{\partial x_{i}}\frac{\mathrm{d} x_i}{\mathrm{d} t}= \sum_{i=1}^{n} \frac{\partial B}{\partial x_{i}}f_i(x, g(x))
\end{equation}
where $f_i$ is the $i$-th component of $f$ in (\ref{eqn:dynamics}). It measures the rate of change of $B$ over time along the direction of the system dynamics. When the vector field is clear in the context, we can also write $\dot B(x)$ instead of $\lie B(x)$. 
\end{definition}
\begin{definition}[Barrier Functions~\cite{prajna2007framework,ames2019control}] \label{def:barrier_func}
Consider a dynamical system $\dot{x} = f$ in state space $X \subseteq \mathbb{R}^n$. Define $X_s\subseteq X$ as the safe region and $X_u$ as the unsafe region, with $X_s\cap X_u=\emptyset$ and $X_s \cup X_u = X$. Let $B(x): \mathbb{R}^n \rightarrow \mathbb{R}$ be a continuous function. We say $B$ is a barrier function for the system with respect to $(X_s,X_u)$ if:
\begin{equation} \label{equ:barrier_function}
\begin{aligned}
    B(x) &\geq 0, \forall x \in X_s\\
    B(x) &< 0, \forall x \in X_u \\
    \lie B(x)&> 0, \forall x \text{ such that } B(x) = 0
\end{aligned}
\end{equation}
When these conditions are met, we call the zero-level set of $B(x)$ the {\em safety barrier} of the system with respect to $(X_s,X_u)$. 
\end{definition}
The Lie derivative condition is often replaced by a smooth version~\cite{ames2014control} $\lie B(x)\geq - \fixed{\alpha}{\gamma} (B(x))$ for an extended class-$\mathcal{K}_{\infty}$ function. 
We choose $\alpha(B(x))$ as a linear function $\gamma B(x)$ with some positive parameter $\gamma$. This condition reduces to the standard requirement on the Lie derivatives at the boundary, and poses a stronger requirement at other states that can be tuned by $\gamma$. The benefit of using this condition is that it avoids solving the equation $B(x)=0$ and can be easily turned into a loss function for learning-based approaches. Once the states near the barrier satisfies this condition, the states inside it would form a forward invariant set, and any trajectory starting from inside the safety set will never cross the safety barrier. 

\paragraph{Kinematic, Dynamic, and Black-box Vehicle Models} 
The commonly used kinematic vehicle does not consider inertial effects or lateral slip and is only usable at low-speed. The $x,y$ coordinates and yaw angle $\theta$ are the state variables with dynamics: $\dot{x} = v\cos(\theta), \dot{y} = v\sin(\theta),\dot{\theta}=v \tan(\delta)/L, \dot{v} = a$, where the acceleration $a$ and steering input $\delta$ are actions given to the controller, $v$ is the longitudinal speed of the vehicle, and $L$ is the wheel base of the vehicle. 
Almost all existing safe path-tracking control methods are analyzed with this simple model. We will show that the model is also simple for learning and certifying safe neural controllers. We focus more on the highly nonlinear cases 
when the speed is high or the road condition is challenging, the no-slip assumption of wheels are no longer valid. In this case,we need to consider the lateral slip and inertial effects. An illustration of the dynamic model is in Fig.~\ref{fig:bicycle_models_and_flow}. They make the control problem highly challenging. The velocities of the vehicle are modelled by: $\dot{x} = v_x \cos(\theta)  - v_y \sin(\theta)$, $
\dot{y} = v_x \sin(\theta)  + v_y \cos(\theta)$, and $\dot{\theta} = r$. The accelerations are $\dot{v}_x = a - \frac{1}{m} F_{fy} \sin(\delta) + v_y r$, $
    \dot{v}_y = \frac{1}{m} ( F_{fy} \cos(\delta) + F_{fr}) - v_x r$, and $\dot{r} = \frac{1}{I_z} ( l_f F_{fy} \cos(\delta) - l_r F_{ry})$. Here, $\delta$ is the steering input; $v_x, v_y$ and $r$ are longitudinal speed, lateral speed and yaw rate, respectively. 

$l_f$ and $l_r$ are distances from the center of gravity to the front and rear axles. $m$ and $I_z$ are the mass and yaw inertia of the vehicle, respectively. Applying linear approximation for tire cornering stiffness assumption, the lateral forces are $F_{i,y} = -c_i \alpha_i$ where $i \in \{f, r\}$, $c_f$ and $c_r$ are coefficients for the linear approximation of cornering stiffness parameters of the front and rear tires. $\alpha_f$ and $\alpha_r$ are tire slip angles, with
$\alpha_f = \tan^{-1} ( (v_y + l_f r)/v_x) - \delta$ and $\alpha_r = \tan^{-1}((v_y - l_r r)/v_x)$.
For both kinematic and dynamic models, distance error $d_e$ and angle error $\theta_e$ can be derived from $x, y, \theta$ and the path geometry.

\begin{figure*}[h!t]
\centering
\begin{subfigure}{.245\textwidth}
  \centering
  \includegraphics[width=\linewidth]{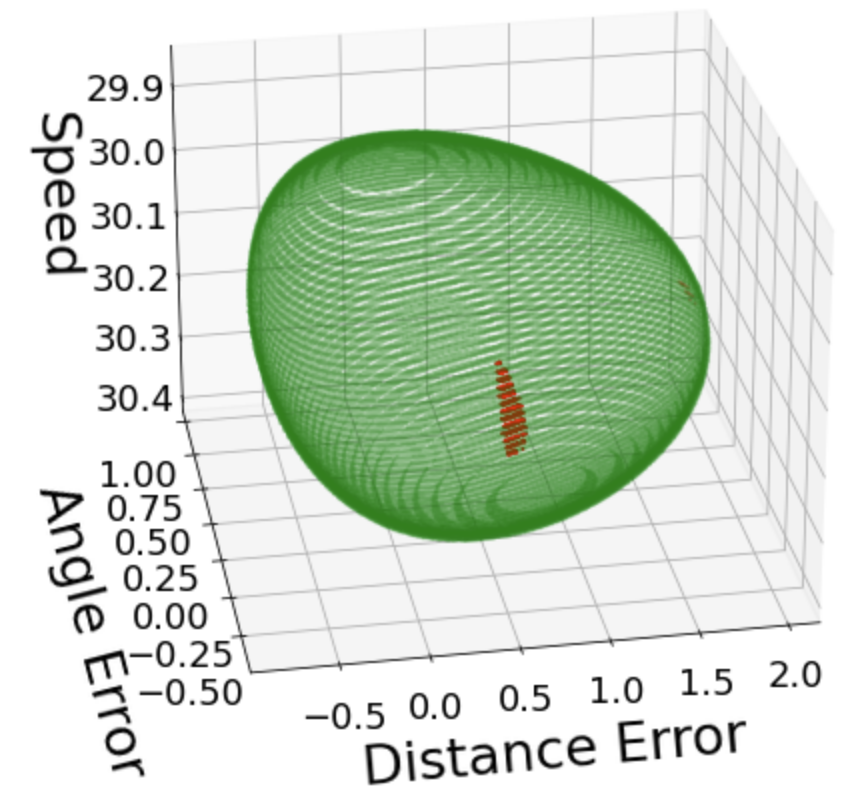} 
  \caption{}
  \label{fig:kinematic_tar_30}
\end{subfigure}
\begin{subfigure}{.245\textwidth}
  \centering
  \includegraphics[width=\linewidth]{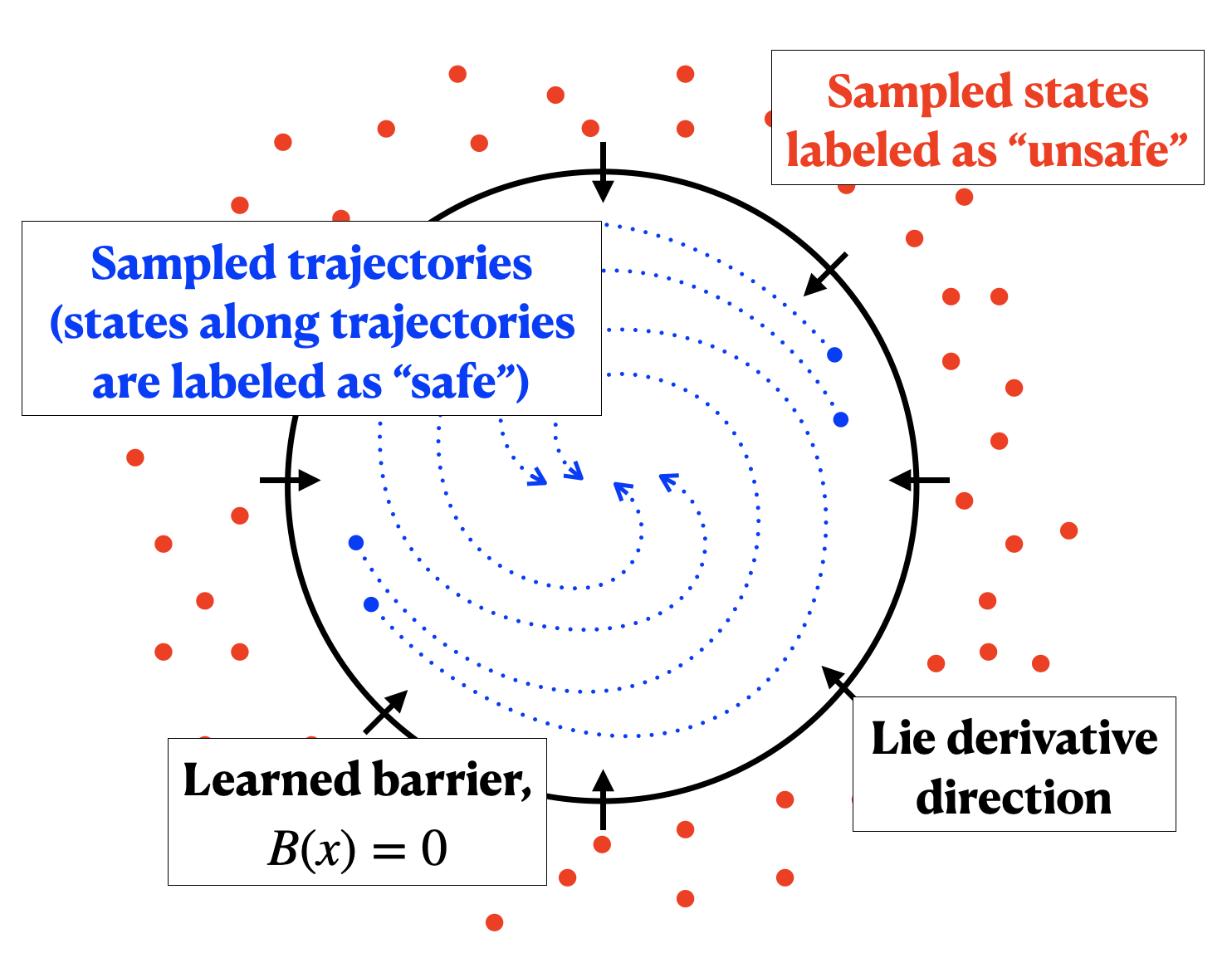} 
  \caption{}
  \label{fig:safe-unsafe}
\end{subfigure}
\begin{subfigure}{.245\textwidth}
  \centering
  \includegraphics[width=\linewidth]{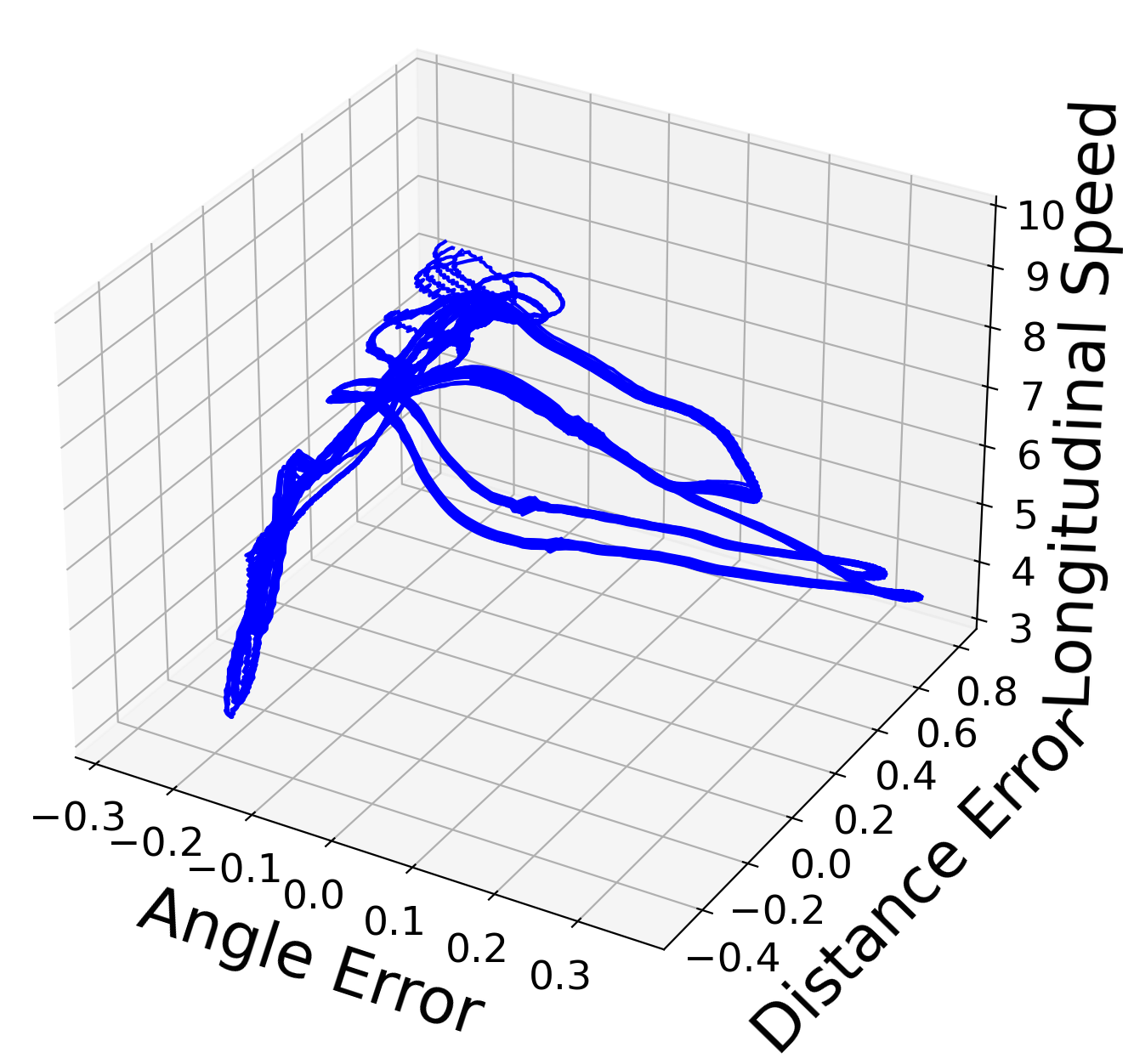} 
  \caption{}
  \label{fig:dynamic_model_trajectories}
\end{subfigure}
\begin{subfigure}{.245\textwidth}
  \centering
  \includegraphics[width=\linewidth]{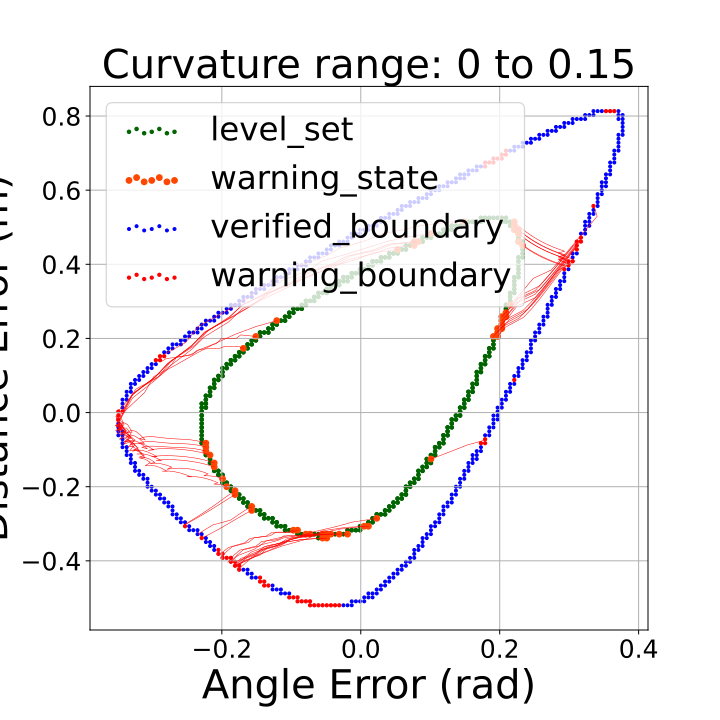}  
  \caption{}
  \label{fig:monitor_0.15}
\end{subfigure}
\caption{\textbf{(a)} Full-state barrier for the kinematic model. The green ball forms the barrier and the red dots shows the small region of uncertified states. \textbf{(b)} Illustration of data used to train the neural barrier. \textbf{(c)} Trajectories of neural controller in the dynamic model in selected three dimensions (angle, distance error and longitudinal speed). \textbf{(d)} Illustration of the safety monitor on a level set inside the safe region (inner curve), the red segments show states that could potentially cross the safety boundary in high curvature environments, while states in the green regions are guaranteed safe within the time window.}
\vspace{-.5cm}
\label{fig:3}
\end{figure*}

\section{Learning Neural Almost-Barrier Functions} \label{sec:learn_barrier_function}
We say a function is an {\em almost-barrier} function if it satisfies the conditions in Definition~\ref{def:almost_barrier} for almost all states other than a small region of the zero-level set. Formally we define:
\begin{definition}[$\varepsilon$-Barrier Functions] \label{def:almost_barrier}
Let $B$ be a scalar function over the state space $X$ of the system in (1). Write the zero-level set of $B$ as $\mathcal{B}(0)=\{x\in X:B(x)=0\}$, and let $\varepsilon\in [0,1)$ be a parameter. Suppose there exists a measurable $\mathcal{C}\subseteq X$ and the ratio of the areas between $\mathcal{C}\cap \mathcal{B}(0)$ and $\mathcal{B}(0)$ is less than $\varepsilon$. 
Namely, $S(\mathcal{C}\cap\mathcal{B}(0))/S(\mathcal{B}(0))\leq \varepsilon$, where $S(A)=\iint_{A}\mathrm{d}S$ indicates surface areas. 
We say $B$ is an $\varepsilon$-barrier function with respect to $(X_s, X_u)$, if $B(x)\geq 0$ for all $x\in X_s$, $B<0$ for all $x\in X_u$, and $\lie B>0$ for all $x\in {\mathcal{C}^c}\cap \mathcal{B}(0)$, where $\mathcal C^c=X\setminus \mathcal C$. 
\end{definition}
In other words, an almost-barrier function satisfies the standard barrier function conditions in Definition~\ref{def:barrier_func} almost everywhere, except in the region of $\mathcal{C}\cap \mathcal{B}(0)$. Clearly, this relaxed notion of barrier functions no longer implies safety in the standard sense. There is now a ``leaking" area of the zero-level set (i.e. $C\cap \mathcal{B}(0)$) from which the system may cross over from the safe area $X_s$ to  the unsafe area $X_u$. For example, Fig.~\ref{fig:kinematic_tar_30} illustrates such "leaking" area as the red region of the barrier function trained for kinematic model. However, finding an $\varepsilon$-barrier is valuable as a quantitative metric for analyzing safety for reasonably small $\varepsilon$. When $\mathcal{C}$ can be explicitly constructed, we can understand the regions of the state space where unsafe behavior may happen, and avoid system operations close to that region. 
In fact, we can perform backward reachability analysis from $\mathcal{C}$ and construct online safety monitor that can alert human operators ahead-of-time before any state in $\mathcal{C}$ is encountered. Fig.3(d) shows such an example, illustrating the almost-barrier and a time-parameterized runtime monitor as the inner boundary. The details will be described in Section~\ref{sec:certify_barrier_function}-(c). 

We now describe the details of the learning procedures. The main steps are: generating safe and unsafe samples, sampling-based estimation of Lie derivatives under full and partial observability, and training the barrier function. The overall pipeline is shown in Fig.~\ref{fig:bicycle_models_and_flow}.

\paragraph{Training control policy and sampling safe and unsafe states} \label{par:sampling_safe_unsafe} We use the standard policy optimization algorithms Soft Actor Critic~\cite{DBLP:conf/icml/HaarnojaZAL18} to learn a control policy $g(\cdot): X\rightarrow \mathbb{R}^m$ using a reward functions that award goal reaching and penalize distance and angle errors. After the training of neural policy converges, the vehicle should show well-controlled behaviors and finish laps on the track without going off the path. We then collect a set of vehicle trajectories from randomized initial positions around the path center, and the states that are visited in these trajectories form the the safe set $X_s$. 

After collecting safe states samples, we generate unsafe state samples $X_u$ around the safe samples with the help of unsupervised learning method $k$-Nearest Neighbor ($k$NN). At each iteration of the data generation process, we uniformly sample $M$ observations in the state space, and add them into a temporary unlabeled candidate set $X_c$. To label each candidate state $x_c \in X_c$, we find its $k$ nearest neighbors inside $X_s \cup X_u \cup X_c$. If \fixed{the majority of the}{more than $k/2$} neighbors are from the safe set, $X_s$, we remove $x_c$ from $X_c$. 
After going through all candidates in $X_c$ in this way, we enlarge the unsafe set $X_u$ with all states that are still in $X_c$. We repeat this procedure until we have enough unsafe samples in $X_u$. The samples generated would not cover the whole state space, but they are sufficient when they can cover the shape of the barrier that separates the safe and unsafe region, as well as covering the barrier region. An illustration of the collected safe and unsafe samples is shown in Fig.~\ref{fig:safe-unsafe}

\paragraph{Training neural barriers} After labeling the safe and unsafe states, we train the barrier function using the following loss function, similar to~\cite{DBLP:conf/cdc/RobeyHLZDTM20,DBLP:journals/fac/ZhaoZCLW21,2020arXiv200608465J}.
{\begin{multline}
\label{eqn:loss_func}
    L(\theta) = w_s \frac{1}{N_s} \sum_{i=0}^{N_s} \phi( - B(x_s^i)) + w_u \frac{1}{N_u} \sum_{i=0}^{N_u} \phi(B(x_u^i)) \\ 
    + w_l\frac{1}{N_s} \sum_{i=0}^{N_s} \phi(- \lie B(x_s^i) - \gamma B(x_s^i)))
\end{multline}}where $\phi(x)=\max(x,0)$, and $w_s, w_u, w_l, \gamma$ are positive parameters for balancing the weights of the different components of the loss. The three terms in $L(\theta)$ correspond to the satisfaction of the three standard conditions for establishing barrier function. For instance, the first term measures the average values of safe states with negative barrier function outputs, and it reaches zero only when the barrier function value is positive on all sampled ``safe" states. Overall, $L(\theta)$ is always non-negative, and it reaches global minimum $L(\theta)=0$ if and only if the barrier conditions in Equation~\ref{equ:barrier_function} are completely satisfied. In evaluating $L(\theta)$, we do not assume the system dynamics $f$ is known. Instead, we approximate the Lie derivative $\lie B$ along sampled trajectories with $\dlie{{B}(s)}=(B(s')-B(s))/\Delta t$, where $s$ and $s'$ are two consecutive states and $\Delta t$ is the time difference between them. Thus, the methods can be used with only black-box access to simulators. 

Note that the loss function only serves as a heuristic for finding candidate barrier functions. We rely on the certification steps in the next section to establish formal guarantees on regions where the barrier conditions are fully satisfied.

\paragraph{Learning barriers under partial observability}
 \label{sec:find_dynamics}
We observe the need of training a lower-dimensional barrier function, over partially observed states when the system state space is high-dimensional. For instance, in the dynamics model with lateral slip, when a sharp corner is encountered with high speed, it is necessary for the vehicle to decelerate in order not to deviate from the track and crash. The corresponding trajectory of the vehicle thus forms a highly non-convex shape. Fig.~\ref{fig:dynamic_model_trajectories} illustrates such a case projected in three dimensions (angle and distance error, and longitudinal speed).
It is very hard to train a neural network to precisely capture the safety boundary of this shape. In addition, with high-dimensions the certification procedures will suffer from curse of dimensionality. Thus, it is important to design methods for capturing the barrier in low-dimensional projections of the states. 

The main difficulty for training low-dimensional barriers under partial observability is the lack of full-state information of the observations that are not seen in the collected trajectories, for approximating the dynamics and Lie derivatives in the loss function in~(\ref{eqn:loss_func}). By further examining longitudinal speed in fig.~\ref{fig:dynamic_model_trajectories},  as well as lateral speed and yaw rate, we observe that at the boundaries where distance and angle errors are large, the other dimensions vary in a very small range. Thus, we can use nearest neighbors to find the dynamics in the unobserved dimensions. Formally, suppose the full states of the environment is $x \in \mathbb{R}^n$ and we aim to train a barrier with partial observation $o \in \mathbb{R}^m$, where $m<n$. After collecting trajectories with full states we select the observation dimensions we want to train a \fixed{barrier function}{CBF} on and form an observation set $O_t$ in $\mathbb{R}^m$. Then, we can learn the system dynamics for a boundary state $o_b$ in the observation space $\mathbb{R}^m$ by finding its nearest neighbor $o_{nn}$ in $O_t$ (i.e. $o_{nn}=\argmin_{o \in O_t} || o -  o_b ||_2$). We then find the nearest neighbor $o_{nn}$'s corresponding full state in $\mathbb{R}^n$ and replace the observation dimensions with the one from the boundary observation. Using the synthesized full state, we can then again simulate the system dynamics of the boundary state.

\section{Certifying Almost-Barrier Functions} \label{sec:certify_barrier_function}
The learning procedures rely on sampled states and approximation of the Lie derivatives. To formally establish almost-barrier conditions, we need to ensure that the zero-level set of learned barrier function correctly separates the safe and unsafe sets, with positive Lie derivatives along the barrier.
In particular, during the certification steps we typically observe many new states where the barrier conditions are violated. We then form an adversarial retraining loop by iteratively adding new states to the training set to improve the barrier function candidate, until the training and certification loop converges.

\begin{table*}[h!t!]
\centering
\caption{Summary of experiment setups and results. \textbf{Lateral}: whether the environment dynamics considers lateral slip; \textbf{Reset}: whether the environment allows arbitrary reset to a given state; \textbf{Partial}: whether the barrier function is trained over partial observation; \bm{$N_s$} and $\bm{N_u}$: Number of safe and unsafe states; \textbf{\# of Grid Cells}: number of grid cells for certification and finding violations; \textbf{Time per Training Round}: Average time used for each round of training; \textbf{\# of Retraining}: the number of retraining rounds for the reported results; \textbf{Certified}: the percentage of verified boundary regions over the whole boundary.}
\label{tab:exp_table}
\begin{tabular}{c|cc|ccccccc}
          & \multicolumn{2}{c|}{Environments} & \multicolumn{7}{c}{Experiments}                                                                  \\ \cline{2-10} 
          & Lateral          & Reset          & Partial & $N_s$   & $N_u$   & \# of Grid Cells & Time per Training Round & \# of Retraining & Certified\% ($1-\varepsilon$) \\ \hline
Kinematic & No               & Yes            & No      & $10000$ & $10000$ & $742500$ (3D)      & $61.50s$                 & $50$          & $99.05\%$ \\ \hline
Dynamic   & Yes              & Yes            & Yes     & $20000$ & $20000$ & $22500$ (2D)      & $51.20s$                & $440$         & $86.4\%$  \\ \hline
TORCS     & Yes              & No             & Yes     & $20000$ & $20000$ & $22500$ (2D)      & $45.05s$                & $587$         & $91.14\%$
\end{tabular}\vspace{-.5cm}
\end{table*}

\paragraph{Robustness analysis of the neural barrier functions} \label{sec:network_robustness}

To certify a learned neural barrier function, we need to certify the barrier conditions in Equation~\ref{equ:barrier_function}.
The first two conditions involve only the value of the learned neural barrier function itself, and are easy to verify. The third condition requires calculating the Lie derivative around $\mathcal{B}(0)$, which is much harder. 
To certify the Lie derivative condition, we need to verify that all states close to the boundary $\mathcal{B}(0)$ satisfy  $\lie B(x)>0$. This is a stronger condition than the original condition but it is necessary to avoid solving the zero-crossing problem. 
Thus, we form a cover of $\mathcal{B}(0)$ by gridding the space around the boundary. We could set $\delta$ as half cell width, and the cells with interval bounds $B_L<0$ and $B_U>0$ are zero-crossing.
We could verify if $\lie{B}(x) > 0$ for such cells. Given a cell center $s$ and its local neighborhood $\|x-s\|_{\infty} \leq \delta$, if the lower bound of the Lie derivative is positive, then the barrier condition is satisfied. 
From Equation~\ref{eqn:lie}, we need to bound the product of the partial derivative of the neural barrier function, ${\partial B(x)}/{\partial x_i}$, and the system dynamics $f_i(x, g(x))$. 
First, the partial derivatives of $B(x)$ need to be bounded through robustness analysis of neural networks. Suppose we parameterize $B(x)$ as a neural network with one hidden-layer and tanh activation $B(x) = W_{1,:}^{(2)} \sigma(W^{(1)}x+b^{(1)}) + b_1^{(2)}.$ Through differentiation, the partial derivative of the neural barrier function is
$
    W_{1,:}^{(2)} \left[ \sigma^\prime(W^{(1)}x+b^{(1)}) \odot W_{:,i}^{(1)} \right],
$
where $\odot$ is the element-wise multiplication and $\sigma^\prime(\cdot)$ is the derivative of tanh function. 
For the local neighborhood of $s$, we write the interval bounds as $l_i^{(2)} \leq \frac{\partial B(x)}{\partial x_i} \leq u_i^{(2)}$
We can then derive $u_i^{(2)}$ and $l_i^{(2)}$ via Holder's inequality. 
Thus, we can propagate the interval bounds on the input $x$ and obtain bounds on the partial derivatives of $\frac{\partial B}{\partial x}$. Note that this calculation can be repeated for each layer to work for neural networks with more layers. 

To bound the system dynamics range in each grid cell, we estimate the Jacobian of the dynamics in the local neighborhood through sampling the dynamics of states that are an $\delta$ away from the grid center $s$ in each dimension. Let $a$ be a sampled sate, and we write the hyperbox around $a$ as $[a]_{\delta}$. 
Using interval arithmetic, we have $L_f B([a]_{\delta})\subseteq \nabla B([a]_{\delta})\circ f([a]_{\delta})$, where $\circ$ is interval dot product.
First, we estimate the value of $f(a)$ through finite difference over time $f(x)\approx (x(t+\Delta t)-x(t))/\Delta t$ evaluated at $x(t)=a$.  
where we assume $\ddot x(t+\lambda \Delta t)\Delta t$ is tiny, $\lambda\in[0,1]$. 
That is, we can assume $\|\ddot{x}(t+\lambda \Delta t)\|\leq L$ and bloat the range around $f(a)$ negligibly by $L\Delta t$. 
We use finite differences to estimate the Jacobian. Writing $f=(f_1,...,f_n)$, with $\delta\rightarrow 0$, we approximate
${\partial f_j}/{\partial x_i}$ at $x=a$ with $ {(f_j(a+\delta e_i)-f_j(a))}/{\delta}$, where $e_i$ is the unit vector in the $x_i$ coordinate.
Thus by taking $n$ (the number of dimensions) samples of $f(x+\delta e_i)$, we can get the estimate of the Jacobian of $f$, $J_f(a)$ (each sample used to estimate all functions in $f$), where $e_i$ is the unit vector in $x_i$, and again assuming that the second-order derivatives are negligible by slightly bloating of the estimated range. 

\paragraph{Retraining the Barrier Using Violation States} To further improve our neural barrier function, we can retrain it by adding the states that violate barrier conditions during certification. 
We say a state $x$ is a counterexample if $B_U(x) > 0$, $B_L(x) < 0$, and $\lie B(x) < 0$. This means that the state is around the safety boundary yet its Lie derivative is in the wrong direction. After collecting these states, we fix the labels on them by applying similar $k$NN approach as in Section~\ref{sec:learn_barrier_function}-(a): we label the state as either "safe" or "unsafe" depending the majority vote of its nearest neighbors. We then add these to the original training set and learn a new candidate barrier function. These retraining steps have significant effects on the shape of the learned barrier as illustrated in~Fig.~\ref{fig:retraining}, which will be explained in the evaluation Section~\ref{sec:exp}-(c). 

\paragraph{Generating Safety Monitors from Almost-Barrier Functions} When the adversarial training loop converges, we can pinpoint small regions on the boundary of almost-barrier functions where the Lie derivative conditions are not certified. 
Thus, we can construct runtime monitor that incorporates reachability analysis to identify potentially unsafe states so that human operators may take over ahead of time to avoid unsafe operations. 
Fig.~\ref{fig:monitor_0.15} illustrates, on a level set inside the safe region, states that may reach the uncertified region of the safety boundary within 50 time steps under worst-case curvature conditions.

\section{Evaluation Results} \label{sec:exp}

We demonstrate the learned and certified barrier on three different environments. In addition, we evaluate three controllers in the context of the trained safety barrier trained with neural policy, and we demonstrate the importance of re-sampling and retraining in our training pipeline.

We evaluate the almost-barrier function training and verification pipeline in three simulation environments: (a) kinematic bicycle model with no-slip assumption, (b) dynamics model with tire-slip lateral control, and (c) TORCS racing environment, which is a black-box high-fidelity vehicle dynamics simulator. All three environments include different tracks for training and evaluation, and obstacles or vehicle overtaking are not considered.
In (a), the kinematic model contains three state dimensions: distance error, angle error and longitudinal speed. In (b), the dynamics model contains two additional state dimensions: lateral speed and vehicle yaw (spin) rate. In (c), the state information for TORCS is more complex and can be found in \cite{torcs_environment}. 
In the kinematic and dynamic models, we have the ability to arbitrarily sample the vehicle states (arbitrary reset), while in TORCS we are limited to sample a small range of initial configurations which poses additional difficulty for the sampling procedures. The action space in the three environments is the same. 
For the neural control policy, we use a two-layer fully connected neural network, with 128 hidden nodes in each layer and ReLU activations. The neural network for the barrier function has a single layer, with 256 hidden nodes and $tanh$ activation function. 
The distance error measures how far the vehicle deviates from the track center, and the angle error measures the difference between the forward direction of the track and the heading direction of the vehicle. 
Table~\ref{tab:exp_table} shows a summary of the experiment setups and results.

\begin{figure}[ht]
\centering
\begin{subfigure}{.23\textwidth}
  \centering
  \includegraphics[width=\linewidth]{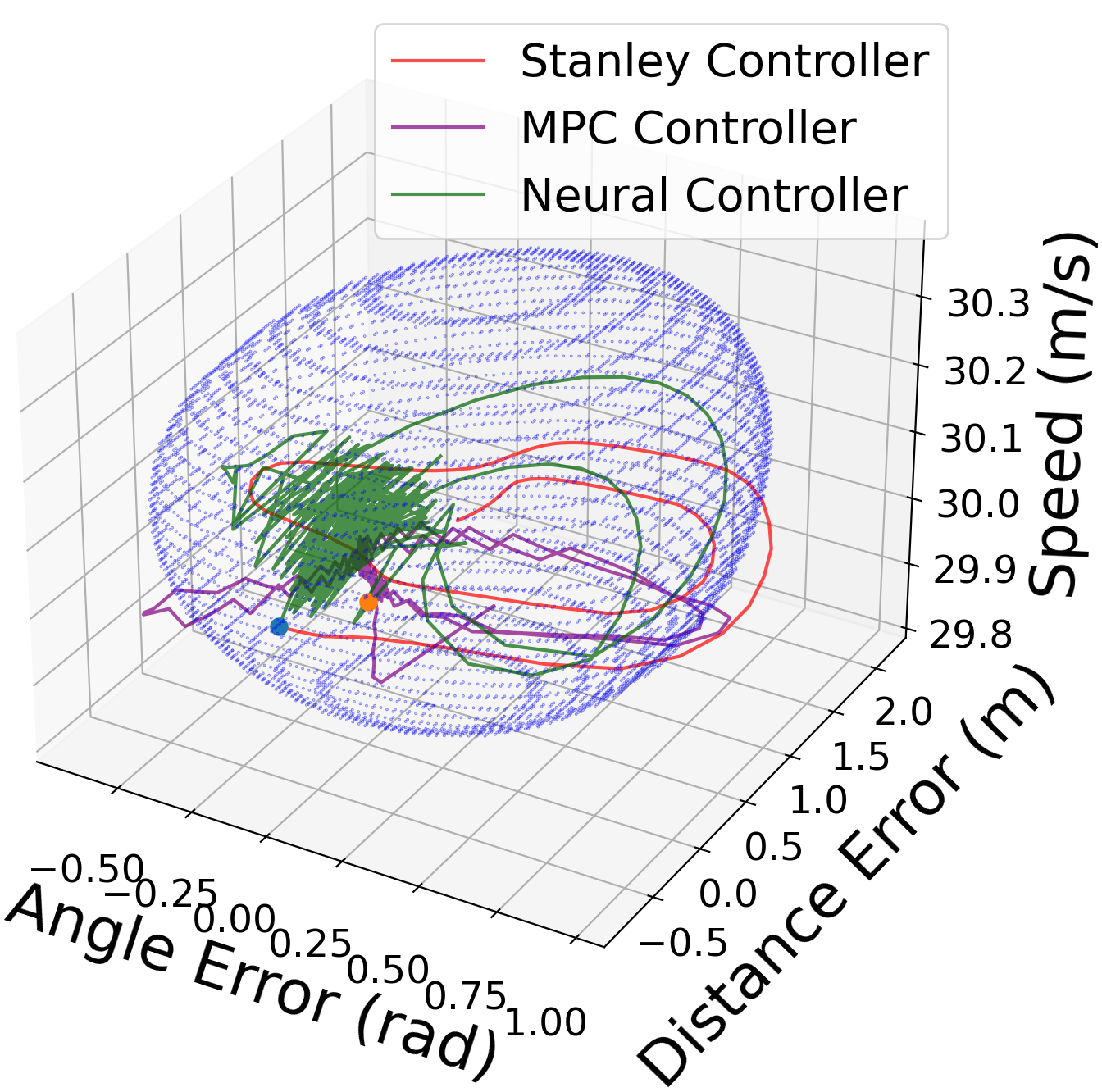} 
  \caption{}
  \label{fig:comparison_kinematic_cbf}
\end{subfigure}
\begin{subfigure}{.23\textwidth}
  \centering
  \includegraphics[width=\linewidth]{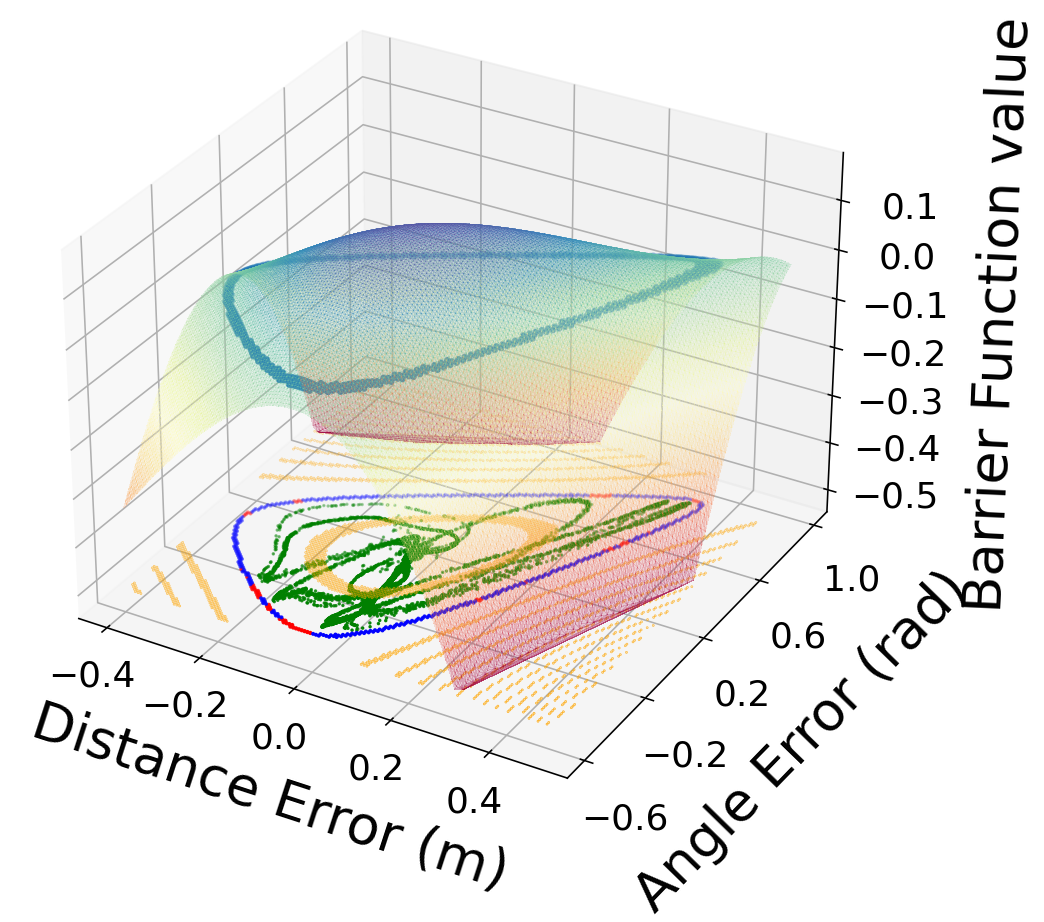}  
  \caption{}
  \label{fig:cbf_2d_dynamic}
\end{subfigure}
\begin{subfigure}{.23\textwidth}
  \centering
  \includegraphics[width=\linewidth]{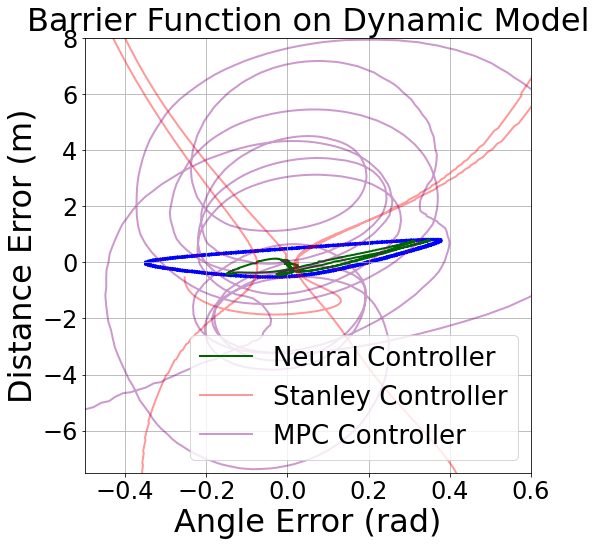}  
  \caption{}
  \label{fig:comparison_dynamic_cbf}
\end{subfigure}
\begin{subfigure}{.23\textwidth}
  \centering
  \includegraphics[width=\linewidth]{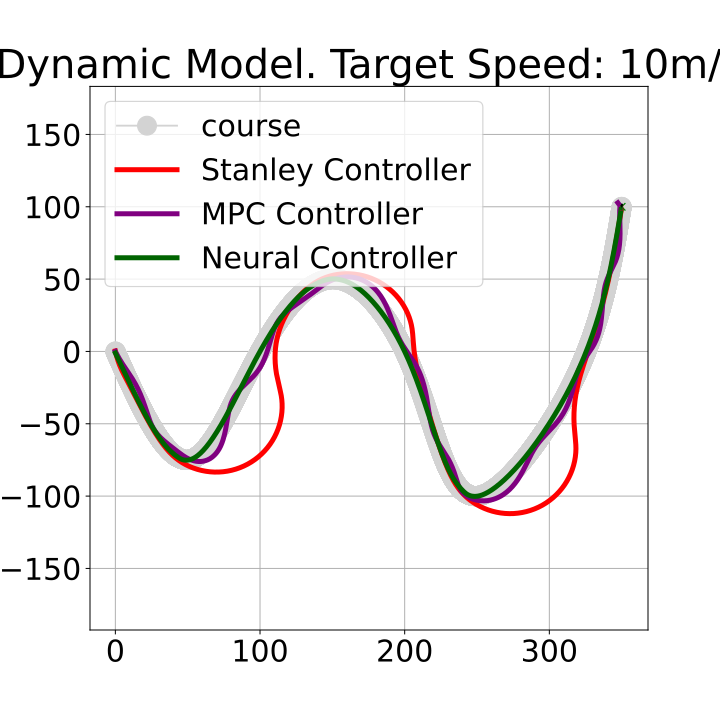}
  \caption{}
  \label{fig:comparison_dynamic_path}
\end{subfigure}
\caption{\textbf{(a)} Comparing Stanley  (red),  MPC  (purple)  and  neural controller  (green)  in  the  kinematic  model, illustrated  with respect to the learned  neural  barrier  (blue). \textbf{(b)} Learned barriers for the dynamics model. The  surface  plot  shows  the  value  of  learned barrier function, and the x-y plane shows the level sets, with vehicle trajectories within the safe set. The zero-level set is colored in  red and blue. \textbf{(c)} 2D projections of the vehicle trajectories under the three controllers in the dynamic model environment, where only the neural controller stays within the barrier. \textbf{(d)} Tracking performance of the three controllers in the dynamic model environment.}
\label{fig:controller_comparison}
\vspace{-.3cm}
\end{figure}

\begin{figure}[h]
\centering
\begin{subfigure}{.20\textwidth}
  \centering
  \includegraphics[width=\linewidth]{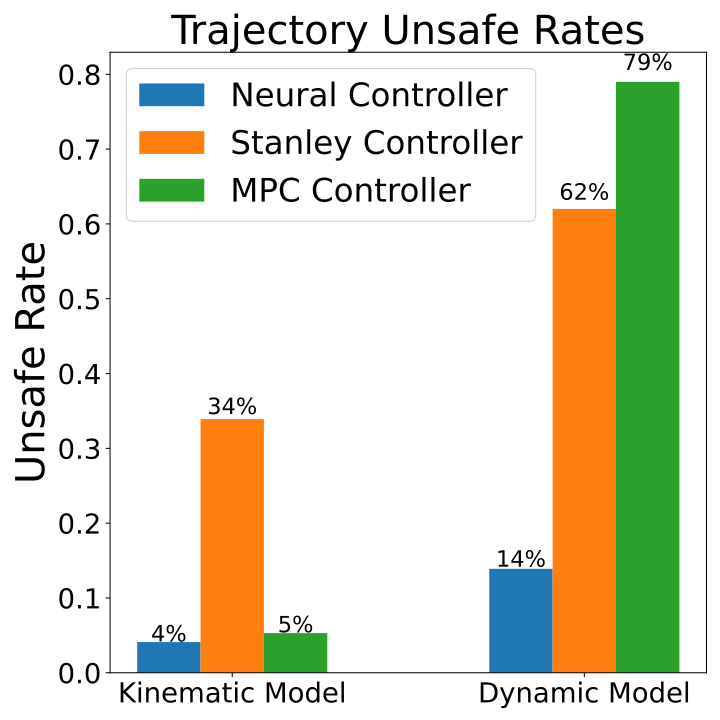}
  \caption{}
  \label{fig:unsafe_stats}
\end{subfigure}
\begin{subfigure}{.27\textwidth}
  \centering
  \includegraphics[width=\linewidth]{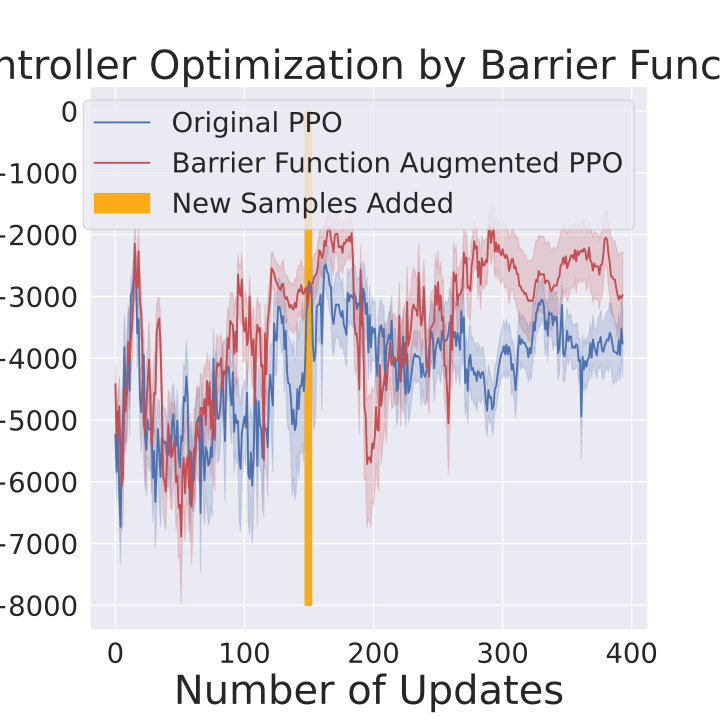}
  \caption{}
  \label{fig:controller_optimization}
\end{subfigure}
\caption{\textbf{(a)} The unsafe trajectory rates of neural, Stanley and MPC controllers, on kinematic and dynamic vehicle model environments. \textbf{(b)} The effect of barrier-function-augmented neural controller training. At the $150$ time step, a barrier function is trained, and the boundary regions are added as the initial states in the reinforcement learning steps onward.}
\label{fig:ral_added}
\vspace{-.3cm}
\end{figure}

\paragraph{Learning and certification results in the kinematic, dynamic, and blackbox models} In the simplest setting, we have a simple bicycle vehicle model without wheel slipping and we can arbitrarily reset the vehicle to any state. Fig.~\ref{fig:kinematic_tar_30} shows the full-state barrier function on the Kinematic Model.
In this experiment, we let the vehicle follow a speed of $30m/s$. 
The green points are where the boundary conditions are satisfied and the red points show uncertified grid cells. The violation rate among the boundary is $0.95\%$, establishing an almost-barrier for a very small $\varepsilon$.

In a more challenging setting of the dynamics model with lateral slip, the vehicle can crash if a hard turn was applied under high speed. 
We train barrier functions with partial observation in the 2D space of distance and angular errors. We can sample arbitrary states in this simulation for estimating the system dynamics. 
In the dynamic model, we let the vehicle follow a target speed of $10m/s$. In Fig.~\ref{fig:cbf_2d_dynamic}, we show the value of the barrier function on the two observation dimensions, with level sets plotted in the x-y plane. The final barrier separates the safe trajectories from the unsafe observations, and $86.4\%$ boundary states satisfy the Lie derivative requirements.

The third environment we consider is the TORCS racing environment, which models more realistic slipping effects under high speeds. 
We let the vehicle follow a speed of $20m/s$ and train a neural barrier function on TORCS using partial observation with angle and distance errors. Fig.~\ref{fig:cbf_2d_torcs} shows the barrier function for the TORCS environment, with a violation rate of $8.86\%$. The lack of the ability to reset to arbitrary states makes it much harder to certify.

The above experiments demonstrate the feasibility of training and quantatively certifying barrier functions for neural controllers on nonlinear high-fidelity path tracking environments, which produces barriers with the majority regions guaranteed to satisfy safety conditions.

\paragraph{Consistency of control performance with learned barriers} We further validate that the learned neural barriers certify safety of the neural controllers. In Figure \ref{fig:controller_comparison}, we see that the vehicle trajectories using the neural controller are indeed contained in the safe set defined by the neural barrier. In contrast, we show trajectories using the Stanley and MPC controllers violate the safety barriers. Fig.~\ref{fig:comparison_dynamic_path} show the control behavior of the neural, Stanley and MPC controllers on the evaluation path, for dynamic models. Fig.~\ref{fig:comparison_kinematic_cbf} and~\ref{fig:comparison_dynamic_cbf} compare the trajectories on the dimensions where the barrier functions are defined. For both models, the trajectory from the neural controller never leaves the barrier, while the vehicle trajectories under the other controllers escape the corresponding safe set.

We also quantitatively measure the unsafe trajectories rates. For the kinematic and dynamic vehicle models, we randomly sample initial states inside the safety barrier, and simulate several timesteps forward with the neural, Stanley and MPC controllers. We measure the percentage of trajectories that escape the trained barrier, shown in Figure~\ref{fig:unsafe_stats}. The neural controller has the lowest unsafe rate, consistent with the small portion of the safety barrier that is not fully certified. 

\begin{figure}[t]
\centering
\includegraphics[width=0.22\textwidth]{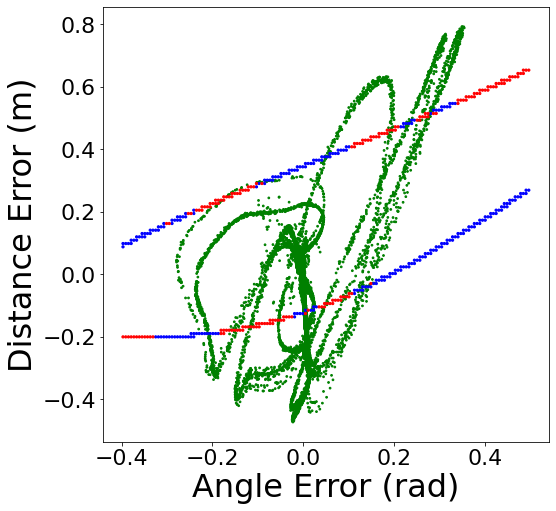}
\includegraphics[width=0.22\textwidth]{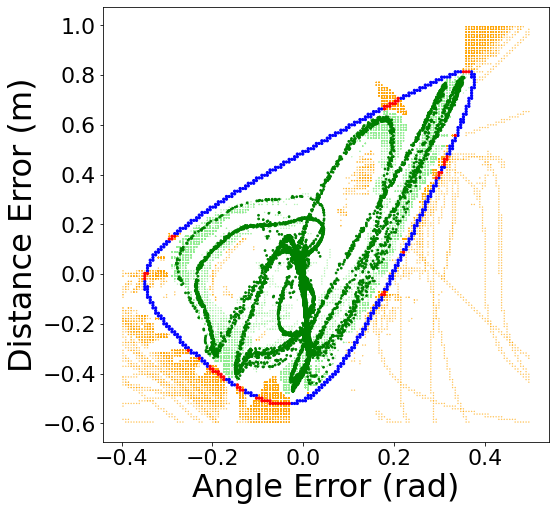}
\caption{Illustration of the importance of adversarial training. Left: initial barrier (blue and red) and original safe data (green); Right: final barrier after continuous training and relabeled violations (safe: light green; unsafe: yellow). }
\vspace{-.5cm}
\label{fig:retraining}
\end{figure} 

\paragraph{Importance of adversarial training} 
The iterative adversarial training steps allow the neural barrier to adjust its shape to improve the satisfaction of the barrier conditions. In Fig.~\ref{fig:retraining}, we show how the barrier for dynamic model changes during continuous training, where we train until convergence, re-label violations as safe or unsafe data, and retrain the barrier function with amended dataset. We consider the adversarial retraining loop to have converged when newly added samples do not further decrease violation. As shown in Fig.~\ref{fig:retraining}, the initial barrier function is not close to forming a valid safety boundary, with a large portion (more than $50\%$) of violations, whereas after the training converges it becomes much closer to a valid safety boundary. 

\paragraph{Augmenting policy training using trained barrier function} When a barrier function is trained and verified, it can provide guidance for policy optimization. The boundary region, especially those that are not fully verified, are more likely the "critical" states that should be trained more on. We can then go back to the policy training to occasionally guide the policy to visit those barrier states (for example, if the environment provides arbitrary resets, start the trajectory from those states as initial states). Figure~\ref{fig:controller_optimization} provides results of PPO training augmented by the trained barrier function. After updating the policy for $150$ times, a barrier function is trained and the boundary states are occasionally used as the initial states for the policy training onward. It can be shown that this procedure results in an improved performance curve compared to the original PPO algorithm.

\section{Conclusion}

We proposed new methods for learning and certifying almost-barrier functions for neural controllers for path-tracking control in self-driving. We described methods for iterative training of the neural barriers and certification methods using robustness analysis of neural networks. 
In experiments with environments ranging from simplistic to high-fidelity simulations, we showed the feasibility of the proposed approach for quantitative safety analysis of neural controllers. Future directions involve better sampling and certification techniques for scaling up. We believe learning neural almost-barriers can be a first step towards ensuring safety of neural control methods in practice, 
with ultimate goals to facilitate safe human-robot interaction for robotic systems with learning-based components in general. 

\bibliographystyle{IEEEtran}
\bibliography{ref}

\begin{thebibliography}{10}
\providecommand{\url}[1]{#1}
\csname url@samestyle\endcsname
\providecommand{\newblock}{\relax}
\providecommand{\bibinfo}[2]{#2}
\providecommand{\BIBentrySTDinterwordspacing}{\spaceskip=0pt\relax}
\providecommand{\BIBentryALTinterwordstretchfactor}{4}
\providecommand{\BIBentryALTinterwordspacing}{\spaceskip=\fontdimen2\font plus
\BIBentryALTinterwordstretchfactor\fontdimen3\font minus
  \fontdimen4\font\relax}
\providecommand{\BIBforeignlanguage}[2]{{%
\expandafter\ifx\csname l@#1\endcsname\relax
\typeout{** WARNING: IEEEtran.bst: No hyphenation pattern has been}%
\typeout{** loaded for the language `#1'. Using the pattern for}%
\typeout{** the default language instead.}%
\else
\language=\csname l@#1\endcsname
\fi
#2}}
\providecommand{\BIBdecl}{\relax}
\BIBdecl

\bibitem{Snider-2009-10165}
J.~M. Snider, ``Automatic steering methods for autonomous automobile path
  tracking,'' Carnegie Mellon University, Pittsburgh, PA, Tech. Rep.
  CMU-RI-TR-09-08, February 2009.

\bibitem{drivingsurvey}
B.~{Paden}, M.~{Čáp}, S.~Z. {Yong}, D.~{Yershov}, and E.~{Frazzoli}, ``A
  survey of motion planning and control techniques for self-driving urban
  vehicles,'' \emph{IEEE Transactions on Intelligent Vehicles}, vol.~1, no.~1,
  pp. 33--55, 2016.

\bibitem{DBLP:conf/rss/KaufmannLR0K020}
E.~Kaufmann, A.~Loquercio, R.~Ranftl, M.~M{\"{u}}ller, V.~Koltun, and
  D.~Scaramuzza, ``Deep drone acrobatics,'' in \emph{Robotics: Science and
  Systems XVI, 2020}, M.~Toussaint, A.~Bicchi, and T.~Hermans, Eds., 2020.

\bibitem{DBLP:conf/rss/PengCZLTL20}
X.~B. Peng, E.~Coumans, T.~Zhang, T.~E. Lee, J.~Tan, and S.~Levine, ``Learning
  agile robotic locomotion skills by imitating animals,'' in \emph{Robotics:
  Science and Systems XVI 2020}, M.~Toussaint, A.~Bicchi, and T.~Hermans, Eds.,
  2020.

\bibitem{DBLP:journals/jfr/ThrunMDSADFGHHLOPPSSDJKMRNJABDEKNM06}
S.~Thrun, M.~Montemerlo, H.~Dahlkamp, D.~Stavens, A.~Aron \emph{et~al.},
  ``Stanley: The robot that won the {DARPA} grand challenge,'' \emph{J. Field
  Robotics}, vol.~23, no.~9, pp. 661--692, 2006.

\bibitem{DBLP:conf/ivs/KongPSB15}
J.~Kong, M.~Pfeiffer, G.~Schildbach, and F.~Borrelli, ``Kinematic and dynamic
  vehicle models for autonomous driving control design,'' in \emph{2015 {IEEE}
  Intelligent Vehicles Symposium, {IV} 2015, Seoul, South Korea, June 28 - July
  1, 2015}, 2015, pp. 1094--1099.

\bibitem{prajna2007framework}
S.~Prajna, A.~Jadbabaie, and G.~J. Pappas, ``A framework for worst-case and
  stochastic safety verification using barrier certificates,'' \emph{IEEE
  Trans. Autom. Control}, vol.~52, no.~8, pp. 1415--1428, 2007.

\bibitem{ames2014control}
A.~D. Ames, J.~W. Grizzle, and P.~Tabuada, ``Control barrier function based
  quadratic programs with application to adaptive cruise control,'' in
  \emph{Proc. Conf. Decis. Control}, Los Angeles, CA,, December 2014, pp.
  6271--6278.

\bibitem{ames2017control}
A.~D. Ames, X.~Xu, J.~W. Grizzle, and P.~Tabuada, ``Control barrier function
  based quadratic programs for safety critical systems,'' \emph{IEEE Trans.
  Autom. Control}, vol.~62, no.~8, pp. 3861--3876, 2017.

\bibitem{weng2018towards}
T.-W. Weng, H.~Zhang, H.~Chen, Z.~Song, C.-J. Hsieh \emph{et~al.}, ``Towards
  fast computation of certified robustness for relu networks,'' \emph{ICML},
  2018.

\bibitem{weng2019proven}
L.~Weng, P.-Y. Chen, L.~Nguyen, M.~Squillante, A.~Boopathy \emph{et~al.},
  ``Proven: Verifying robustness of neural networks with a probabilistic
  approach,'' \emph{ICML}, 2019.

\bibitem{10.1007/11730637_22}
A.~Girard and G.~J. Pappas, ``Verification using simulation,'' in \emph{Hybrid
  Systems: Computation and Control}, J.~P. Hespanha and A.~Tiwari, Eds.\hskip
  1em plus 0.5em minus 0.4em\relax Berlin, Heidelberg: Springer Berlin
  Heidelberg, 2006, pp. 272--286.

\bibitem{epsilon-approximationof}
A.~Puri, P.~Varaiya, and V.~Borkar, ``$esilon$-approximation of differential
  inclusions,'' 1995.

\bibitem{10.1007/3-540-36580-X_5}
E.~Asarin, T.~Dang, and A.~Girard, ``Reachability analysis of nonlinear systems
  using conservative approximation,'' in \emph{Hybrid Systems: Computation and
  Control}, O.~Maler and A.~Pnueli, Eds.\hskip 1em plus 0.5em minus 0.4em\relax
  Berlin, Heidelberg: Springer Berlin Heidelberg, 2003, pp. 20--35.

\bibitem{1656431}
{Zhi Han} and B.~H. {Krogh}, ``Reachability analysis of nonlinear systems using
  trajectory piecewise linearized models,'' in \emph{2006 American Control
  Conference}, 2006, pp. 6 pp.--.

\bibitem{4738704}
M.~{Althoff}, O.~{Stursberg}, and M.~{Buss}, ``Reachability analysis of
  nonlinear systems with uncertain parameters using conservative
  linearization,'' in \emph{2008 47th IEEE Conference on Decision and Control},
  2008, pp. 4042--4048.

\bibitem{DBLP:conf/hybrid/DangMT10}
T.~Dang, O.~Maler, and R.~Testylier, ``Accurate hybridization of nonlinear
  systems,'' in \emph{Proceedings of the 13th {ACM} International Conference on
  Hybrid Systems: Computation and Control, {HSCC} 2010}, K.~H. Johansson and
  W.~Yi, Eds.\hskip 1em plus 0.5em minus 0.4em\relax {ACM}, 2010, pp. 11--20.

\bibitem{1166525}
A.~{Chutinan} and B.~H. {Krogh}, ``Computational techniques for hybrid system
  verification,'' \emph{IEEE Transactions on Automatic Control}, vol.~48,
  no.~1, pp. 64--75, 2003.

\bibitem{1215682}
C.~J. {Tomlin}, I.~{Mitchell}, A.~M. {Bayen}, and M.~{Oishi}, ``Computational
  techniques for the verification of hybrid systems,'' \emph{Proceedings of the
  IEEE}, vol.~91, no.~7, pp. 986--1001, 2003.

\bibitem{8263867}
S.~L. {Herbert}, M.~{Chen}, S.~{Han}, S.~{Bansal}, J.~F. {Fisac}, and C.~J.
  {Tomlin}, ``Fastrack: A modular framework for fast and guaranteed safe motion
  planning,'' in \emph{2017 IEEE 56th Annual Conference on Decision and Control
  (CDC)}, 2017, pp. 1517--1522.

\bibitem{4282788}
G.~M. {Hoffmann}, C.~J. {Tomlin}, M.~{Montemerlo}, and S.~{Thrun}, ``Autonomous
  automobile trajectory tracking for off-road driving: Controller design,
  experimental validation and racing,'' in \emph{2007 American Control
  Conference}, 2007, pp. 2296--2301.

\bibitem{8998094}
V.~{K}, M.~{Ambalal Sheta}, and V.~{Gumtapure}, ``A comparative study of
  stanley, lqr and mpc controllers for path tracking application (adas/ad),''
  in \emph{2019 IEEE International Conference on Intelligent Systems and Green
  Technology (ICISGT)}, 2019, pp. 67--674.

\bibitem{7795743}
S.~{Dominguez}, A.~{Ali}, G.~{Garcia}, and P.~{Martinet}, ``Comparison of
  lateral controllers for autonomous vehicle: Experimental results,'' in
  \emph{2016 IEEE 19th International Conference on Intelligent Transportation
  Systems (ITSC)}, 2016, pp. 1418--1423.

\bibitem{DBLP:conf/icra/SchwartingAPKR17}
W.~Schwarting, J.~Alonso{-}Mora, L.~Pauli, S.~Karaman, and D.~Rus, ``Parallel
  autonomy in automated vehicles: Safe motion generation with minimal
  intervention,'' in \emph{2017 {IEEE} International Conference on Robotics and
  Automation, {ICRA} 2017, Singapore, Singapore, May 29 - June 3, 2017}, 2017,
  pp. 1928--1935.

\bibitem{wieland2007constructive}
P.~Wieland and F.~Allg{\"o}wer, ``Constructive safety using control barrier
  functions,'' in \emph{Proc. IFAC Symp. Nonlin. Control Syst.}, Pretoria,
  South Africa, August 2007, pp. 462--467.

\bibitem{parrilo2000structured}
P.~A. Parrilo, ``Structured semidefinite programs and semialgebraic geometry
  methods in robustness and optimization,'' Ph.D. dissertation, California
  Institute of Technology, 2000.

\bibitem{ames2019control}
A.~D. Ames, S.~Coogan, M.~Egerstedt, G.~Notomista, K.~Sreenath, and P.~Tabuada,
  ``Control barrier functions: Theory and applications,'' in \emph{2019 18th
  European Control Conference (ECC)}, Naples, Italy, June 2019, pp. 3420--3431.

\bibitem{xu2017correctness}
X.~Xu, J.~W. Grizzle, P.~Tabuada, and A.~D. Ames, ``Correctness guarantees for
  the composition of lane keeping and adaptive cruise control,'' \emph{IEEE
  Transactions on Automation Science and Engineering}, vol.~15, no.~3, pp.
  1216--1229, 2017.

\bibitem{wang2018permissive}
L.~Wang, D.~Han, and M.~Egerstedt, ``Permissive barrier certificates for safe
  stabilization using sum-of-squares,'' in \emph{2018 Annual American Control
  Conference (ACC)}.\hskip 1em plus 0.5em minus 0.4em\relax IEEE, 2018, pp.
  585--590.

\bibitem{DBLP:conf/aaai/ChengOMB19}
R.~Cheng, G.~Orosz, R.~M. Murray, and J.~W. Burdick, ``End-to-end safe
  reinforcement learning through barrier functions for safety-critical
  continuous control tasks,'' in \emph{The Thirty-Third {AAAI} Conference on
  Artificial Intelligence, {AAAI} 2019}, 2019, pp. 3387--3395.

\bibitem{DBLP:conf/icml/ChengVOCYB19}
R.~Cheng, A.~Verma, G.~Orosz, S.~Chaudhuri, Y.~Yue, and J.~Burdick, ``Control
  regularization for reduced variance reinforcement learning,'' in
  \emph{International Conference on Machine Learning ({ICML}) 2019}, ser.
  Proceedings of Machine Learning Research, K.~Chaudhuri and R.~Salakhutdinov,
  Eds., vol.~97.\hskip 1em plus 0.5em minus 0.4em\relax {PMLR}, 2019, pp.
  1141--1150.

\bibitem{DBLP:conf/cdc/RobeyHLZDTM20}
A.~Robey, H.~Hu, L.~Lindemann, H.~Zhang, D.~V. Dimarogonas \emph{et~al.},
  ``Learning control barrier functions from expert demonstrations,'' in
  \emph{59th {IEEE} Conference on Decision and Control, {CDC} 2020}.\hskip 1em
  plus 0.5em minus 0.4em\relax {IEEE}, 2020, pp. 3717--3724.

\bibitem{DBLP:journals/fac/ZhaoZCLW21}
H.~Zhao, X.~Zeng, T.~Chen, Z.~Liu, and J.~Woodcock, ``Learning safe neural
  network controllers with barrier certificates,'' \emph{Formal Aspects
  Comput.}, vol.~33, no.~3, pp. 437--455, 2021.

\bibitem{katz2017reluplex}
G.~Katz, C.~Barrett, D.~L. Dill, K.~Julian, and M.~J. Kochenderfer, ``Reluplex:
  An efficient smt solver for verifying deep neural networks,'' in
  \emph{ICCAV}, 2017.

\bibitem{hein2017formal}
M.~Hein and M.~Andriushchenko, ``Formal guarantees on the robustness of a
  classifier against adversarial manipulation,'' in \emph{NIPS}, 2017.

\bibitem{weng2018evaluating}
T.-W. Weng, H.~Zhang, P.-Y. Chen, J.~Yi, D.~Su \emph{et~al.}, ``Evaluating the
  robustness of neural networks: An extreme value theory approach,''
  \emph{ICLR}, 2018.

\bibitem{kolter2017provable}
J.~Z. Kolter and E.~Wong, ``Provable defenses against adversarial examples via
  the convex outer adversarial polytope,'' \emph{ICML}, 2018.

\bibitem{zhang2018crown}
H.~Zhang, T.-W. Weng, P.-Y. Chen, C.-J. Hsieh, and L.~Daniel, ``Efficient
  neural network robustness certification with general activation functions,''
  in \emph{NeurIPS}, 2018.

\bibitem{singh2018fast}
G.~Singh, T.~Gehr, M.~Mirman, M.~P\"{u}schel, and M.~Vechev, ``Fast and
  effective robustness certification,'' in \emph{NeurIPS}, 2018.

\bibitem{wang2018efficient}
S.~Wang, K.~Pei, J.~Whitehouse, J.~Yang, and S.~Jana, ``Efficient formal safety
  analysis of neural networks,'' in \emph{NeurIPS}, 2018.

\bibitem{raghunathan2018semidefinite}
A.~Raghunathan, J.~Steinhardt, and P.~S. Liang, ``Semidefinite relaxations for
  certifying robustness to adversarial examples,'' in \emph{NeurIPS}, 2018, pp.
  10\,877--10\,887.

\bibitem{Boopathy2019cnncert}
A.~Boopathy, T.-W. Weng, P.-Y. Chen, S.~Liu, and L.~Daniel, ``Cnn-cert: An
  efficient framework for certifying robustness of convolutional neural
  networks,'' in \emph{AAAI}, 2019.

\bibitem{ko2019popqorn}
C.-Y. Ko, Z.~Lyu, T.-W. Weng, L.~Daniel, N.~Wong, and D.~Lin, ``Popqorn:
  Quantifying robustness of recurrent neural networks,'' \emph{ICML}, 2019.

\bibitem{DBLP:conf/icml/HaarnojaZAL18}
T.~Haarnoja, A.~Zhou, P.~Abbeel, and S.~Levine, ``Soft actor-critic: Off-policy
  maximum entropy deep reinforcement learning with a stochastic actor,'' in
  \emph{Proceedings of the 35th International Conference on Machine Learning,
  {ICML} 2018, Stockholmsm{\"{a}}ssan, Stockholm, Sweden, July 10-15, 2018},
  2018, pp. 1856--1865.

\bibitem{2020arXiv200608465J}
W.~{Jin}, Z.~{Wang}, Z.~{Yang}, and S.~{Mou}, ``{Neural Certificates for Safe
  Control Policies},'' \emph{arXiv e-prints}, p. arXiv:2006.08465, Jun. 2020.

\bibitem{torcs_environment}
D.~Loiacono, L.~Cardamone, and P.~L. Lanzi, ``Simulated car racing
  championship: Competition software manual,'' \emph{CoRR}, vol. abs/1304.1672,
  2013.

\end{thebibliography}

\end{document}